\documentclass[runningheads]{llncs}
\usepackage[T1]{fontenc}
\usepackage{tikz}
\usetikzlibrary{bayesnet}
\usetikzlibrary{arrows}
\usepackage{color}
\usepackage{caption}
\usepackage{subcaption}

\usepackage{graphicx}
\graphicspath{ {./images/} }
\usepackage{amssymb}
\usepackage{amsmath}
\usepackage{cite}

\begin{document}
\title{Mixtures of Unsupervised Lexicon Classification}
\author{Peratham Wiriyathammabhum\inst{1}\orcidID{0000-0001-5567-3104}} 
\authorrunning{Peratham Wiriyathammabhum}
\institute{
\email{peratham.bkk@gmail.com}}
\maketitle              % typeset the header of the contribution
\begin{abstract}
This paper presents a mixture version of the method-of-moment unsupervised lexicon classification by an incorporation of a Dirichlet process.

\end{abstract}
\section{Introduction}

\subsection{Unsupervised Lexicon Classification (Bayesian Lexicon/BayesLex)}
Lexicon classification utilizes a classification scheme where the classification probabilities are estimated from counting word occurrences, $x$, corresponding to each lexicon for each class, $W_0$ and $W_1$, classes $Y \in \{0, 1\}$, as in Eq. \ref{eq1}. This lexicon classification scheme is useful for filtering documents and is less computationally expensive than many other classification methods. 
\begin{equation} \label{eq1}
    \sum_{i \in W_0} x_i \gtrless \sum_{j \in W_1} x_j,
\end{equation}
where $\gtrless$ denotes a decision rule operator.

Unsupervised lexicon classification \cite{eisenstein2017unsupervised} refers to an unsupervised probabilistic weighting scheme where each word in each lexicon will have a weight value instead of counting as one. The weights are learned from a corpus and fitted as a multinomial Na\"ive Bayes model using a method-of-moment estimator on cross-lexical co-occurrence counts. As in many typical probabilistic models, there are many also assumptions which are equal prior likelihood, lexicon completeness, equal predictiveness of words, and equal coverage.

To elaborate more, we provide some derivations starting from the original paper, let we have a prior probability $P_Y$ for the class label $Y$, a likelihood function $P_{X|Y}$, where $X$ is a bag-of-word input vector. Then, we can have a familiar classification rule from the conditional probability $P(y|x)$ as,
\begin{equation}
    P(y|x) = \frac{ P(x|y) P(y)}{\sum_{y'} P(x|y') P(y')}.
\end{equation}
By this, we can also assume that the costs of misclassification are equal and we can have the minimum Bayes risk classification rule as,
\begin{equation} \label{eq2}
   \log{Pr(Y=0)} + \log{P(x|Y=0)} \gtrless \log{Pr(Y=1)} + \log{P(x|Y=1)}.
\end{equation}
Then, we assume $P(x|y)$ is multinomially distributed since we have a multinomial Na\"ive Bayes model such that we can have,
\begin{equation}
\begin{aligned} \label{eq3}
    \log{P(x|y)}& = \log{P_{multinomial} (x; \theta_y, N)}\\
             &    = K(x) + \sum_{i=1}^{V} x_i \log{Pr(W=i|Y=y;\theta)}\\
            &   = K(x) + \sum_{i=1}^{V} x_i \log{\theta_{y,i}}
\end{aligned}
\end{equation}
,where $\theta_y$ is a class vector, $W$ is a word or token, $V$ is the vocab size, $N$ is the total count $\sum_{i=1}^V x_i$, and $x_i$ is the raw count of word $i \in \{1,2, \dots, V\}$ as in the standard bag-of-word definition. $K(x)$ is some function of $x$ that is constant in $y$.

Then, we have a completeness assumption that the lexicon is complete, that is, a word that is not in any lexicon has identical probabilities under both classes. Furthermore, we have another assumption that every word in the lexicon is equally predictive, such that,
\begin{equation}
  \frac{Pr(W=i|Y=y)}{Pr(W=i|Y=\neg{y})} = \frac{ 1 + \gamma}{1 - \gamma},
\end{equation}
where $\neg{y}$ is the label for the opposite class and $\gamma$ is a predictive parameter of the lexicon. Given both assumptions, we can define,
\begin{equation}
  \theta_{y,i}= \begin{cases} (1+\gamma)\mu_i&, i \in W_y\\
  (1-\gamma)\mu_i &, i \in W_{\neg{y}}\\
  \mu_i&, i \notin W_y \cup W_{\neg{y}} \\
  \end{cases},
\end{equation}
where $\mu_i$ is the base probability independent of labels.

We introduce another assumption of equal prior likelihood that $Pr(Y=0) = Pr(Y=1)$ so that we can cancel out the prior terms in Eq. \ref{eq2}. Then, we have our new classification rule similar to Eq. \ref{eq1} as,
\begin{equation} \label{eq6}
\begin{aligned}
    \log{P(x|Y=0)} &\gtrless \log{P(x|Y=1)}, \\
    \sum_{i\in W_0} x_i \log{\frac{ 1 + \gamma}{1 - \gamma}} &\gtrless \sum_{i\in W_1} x_i \log{\frac{ 1 + \gamma}{1 - \gamma}},  \\
\end{aligned}  
\end{equation}
where $\gamma \in (0,1)$. We omitted other proofs and derivations from the original paper since they are unrelated to our focus except that we can have an objective function to solve for $\gamma$ by using the cross label counts and the method-of-moment estimator as for the cross label counts $c_i$, the count of word $i$ with all the opposite lexicon words,
\begin{equation}
c_i = \sum_{t=1}^T \sum_{j \in W_{\neg{y}}} x_i^{t}x_j^{t},
\end{equation}
where $t \in {1,\dots,T}$ for $T$ documents in the collection and $x^{t}$ is its vector of word counts. We got the objective function as,
\begin{equation}
\begin{aligned}
\min_{\gamma^{(0)}, \gamma^{(1)}} & \frac{1}{2}\sum_{i \in W_0} (c_i - E[c_i])^2 + \frac{1}{2}\sum_{j \in W_1} (c_j - E[c_j])^2 \\ s.t.\  & \mu^{(0)}\dot\gamma^{(0)} -  \mu^{(1)}\dot\gamma^{(1)} = 0 \\
& \forall i \in (W_0 \cup W_1) , \gamma_i \in [0,1).
\end{aligned}
\end{equation}
After the $\gamma$ parameters are fitted. The original paper uses 2 formula variants for prediction rules, one derived from multinomial distribution as a multinomial Na\"ive Bayes in Eq. \ref{eq6} and another as a plug-and-play Dirichlet-compound multinomial \cite{madsen2005modeling}, which can be directly substituted from the multinomial distribution to better model burstiness, since the model only models word presences instead of counts, and is more predictive \cite{pang2002thumbs}. We are going to modify these aforementioned equations, especially the prediction rules which compute probabilities as in Eq. \ref{eq3} and Eq. \ref{eq6}, not the $\gamma$ solver.

The thing is whether the model is a multinomial Na\"ive Bayes (maximum likelihood) or a Dirichlet-compound multinomial Na\"ive Bayes (multinomial with a Dirichlet prior), the model does not handle group clustering \cite{teh2004sharing} (or in other words, mixed membership). The group clustering problem arises as there are grouped data which can be subdivided into a set of groups and each group has clusters of data while there are also components which are shared across groups. Some natural examples of grouped data \cite{teh2006hierarchical} are binary markers (single nucleotide polymorphisms or SNPs) in a localized region of the human genome or text documents. In this paper, our focus in on textual data so we will briefly elaborate only this example that, for a document, given a bag-of-words assumption that the orders of words are exchangeable, it can consist of words from many topics shared across many documents in a collection. We continue this Bayesian lexicon line of work and incorporate Dirichlet processes to see whether topical modeling or mixtures can help in giving weights to lexicon better.

\begin{figure}
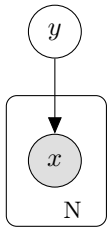
 \label{platenaivebayes}
\begin{center}
  \tikz{
    % nodes
     \node[obs] (x) {$x$};%
     \node[latent,above=of x] (y) {$y$}; %
    % plate
     \plate [inner sep=.3cm,xshift=.02cm,yshift=.2cm] {plate1} {(x)}{N}; %
    % edges
     \edge {y} {x}  }
      \caption{A representation of a Na\"ive Bayes model.}
\end{center}
\end{figure}
\vspace{-1\baselineskip}

\subsection{Mixture Models}
Mixture models \cite{mclachlan1988mixture, mclachlan2019finite} are combinations of basic distributions which can give rises to model complex densities. A mixture model is usually in the form of a linear combination,
\begin{equation}
\begin{aligned}
        f(y) &= \sum_{i=1}^{g} \pi_i f_i(y)\\
        s.t.\  &\sum_{i=1}^{g} \pi_i = 1, \ \pi_i \in [0,1], %\ 0 \le \pi_i \le 1,
\end{aligned}    
\end{equation}
where $f$ denotes a mixture probability mass function, $f_i$ denotes a probability mass function of a base distribution which can also be called as a component of the mixture, $g$ is the number of finite mixtures (or the number of components), and $\pi_i$ is a mixing proportion or a mixing coefficient which also satisfies the requirements to be probabilities. Sometimes, we can view the mixing proportions as prior probabilities of picking the $i$th components. %For many base distributions, like Gaussians, maximizing the log-likelihood is intractable since the log of sums is hard to compute but for count-based distributions like Bernoulli, multinomial, or even priors like Beta or Dirichlet, the log-likelihood is the log of multiples which reduces to a more simple form.

\begin{figure}
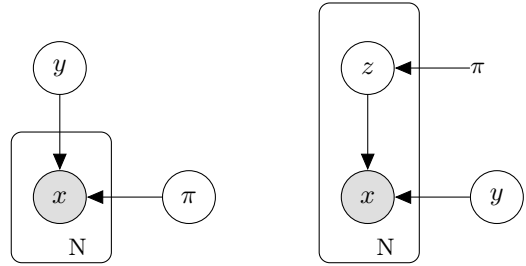
 \label{platemixture}
\begin{center}
\begin{subfigure}[b]{0.3\textwidth}
  \tikz{
    % nodes
     \node[obs] (x) {$x$};%
    \node[latent,right=of x] (p) {$\pi$}; %
     \node[latent,above=of x] (y) {$y$}; %
    % plate
     \plate [inner sep=.3cm,xshift=.02cm,yshift=.2cm] {plate1} {(x) }{N}; %
     % \plate [inner sep=.3cm,xshift=.02cm,yshift=.2cm] {plate1} {(p) }{g}; %
    % edges
     \edge {y} {x}
     \edge {p} {x} }
     \end{subfigure}
     \quad
     \begin{subfigure}[b]{0.3\textwidth}
     \tikz{
    % nodes
     \node[obs] (x) {$x$};%
    \node[latent,above=of x] (z) {$z$}; %
    \node[const,right=of z] (p) {$\pi$}; %
     \node[latent,right=of x] (y) {$y$}; %
    % plate
     \plate [inner sep=.3cm,xshift=.02cm,yshift=.2cm] {plate1} {(x) (z)}{N}; %
     % \plate [inner sep=.3cm,xshift=.02cm,yshift=.2cm] {plate1} {(p) }{g}; %
    % edges
     \edge {y} {x}
     \edge {z} {x}
     \edge {p} {z} }
     \end{subfigure}
      \caption{A representation of a mixture model. The right plate diagram has a latent variable $z$. In the left plate diagram, $\pi$ is a variable instead of a constant.}
      
\end{center}
\end{figure}
\vspace{-1\baselineskip}

\subsection{Dirichlet Processes}
A Dirichlet distribution is a multivariate generalization of a Beta distribution to more than 2 variables, in a similar spirit to a multinomial distribution to a Bernoulli distribution. A $K$-dimensional/outcome Dirichlet distribution, $Dir(\mu|\alpha)$, is defined as a probability density function,
\begin{equation}
    \begin{aligned}
    Dir(\mu|\alpha) &= C^{-1}(\alpha) \prod_{k=1}^{K}\mu^{\alpha_{k}-1}\\
    C(\alpha) &= \frac{\sum_{k=1}^{K} \Gamma(\alpha_k)}{\Gamma(\sum_{k=1}^{K} \alpha_k)}= \frac{\sum_{k=1}^{K} \Gamma(\alpha_k)}{\Gamma(\alpha_0)}\\
    \Gamma(\alpha) &= \int_{0}^{\infty} u^{\alpha-1} e^{-u} du\\
    s.t. \ &\sum_{k=1}^{K} \mu_i = 1, \ \mu_i \in [0,1],\\
    \end{aligned}
\end{equation}
where $\Gamma(\alpha)$ is the gamma function, $C(\alpha)$ is the beta function as a normalizer, all $\alpha_k$ are non-negative parameters of the distribution, $\alpha_o$ is the concentration parameter, and all $\mu_i \in [0,1]$ are non-negative input random base distributions. A symmetric Dirichlet distribution or prior has all $\alpha_k = \frac{\alpha}{K}$.

Dirichlet processes \cite{ferguson1973bayesian, teh2010dirichlet, teh2006hierarchical, teh2004sharing} are discrete-time stochastic processes whose realizations are distributed as Dirichlet distributions. Some people call this as distributions over distributions. Formally, let $(\Theta, B)$ be a measurable space, $G_0$ is a probability measure on this space, and $\alpha_0$ is a positive real constant number. A Dirichlet process $DP(\alpha_0, G_0)$ is the distribution of a random positive measure $G$ over $(\Theta, B)$ such that for any finite measurable partition $(A_1, A_2,\dots,A_g)$ of $\Theta$, a random vector comprising of $G$, $(G(A_1), G(A_2),\dots,G(A_g))$, is distributed as a finite Dirichlet distribution with parameters, $(\alpha_0 G_0(A_1), \alpha_0 G_0(A_2),\dots,\alpha_0 G_0(A_g))$. That is,
\begin{equation}
    \begin{aligned}
        (G(A_1), G(A_2),\dots,G(A_g))\sim Dir(\alpha_0 G_0(A_1), \alpha_0 G_0(A_2),\dots,\alpha_0 G_0(A_g)),
    \end{aligned}
\end{equation}
or, in short, $G \sim DP(\alpha_0, G_0)$. Dirichlet processes can also be defined as a stick-breaking construction, a P\'{o}lya-urn process, a Chinese restaurant process, or taking an infinite limit of finite mixture models. These alternative definitions are too verbose and will not be mentioned in this paper for brevity, except some.

\subsubsection{Dirichlet Process Mixture Models}
One of some interesting applications of Dirichlet processes is Dirichlet process mixture models (DPMMs) by using a Dirichlet process as a prior for the parameters of mixture models. 

\begin{figure}
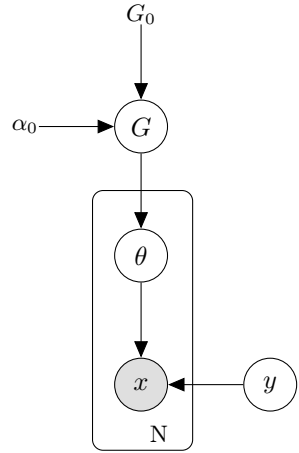
 \label{platedpmixture}
\begin{center}
    \tikz{
        % nodes
         \node[obs] (x) {$x$};%
        \node[latent,above=of x] (t) {$\theta$}; %
         \node[latent,right=of x] (y) {$y$}; %
         \node[latent,above=of t] (g) {$G$}; %
         \node[const,above=of g] (g0) {$G_0$}; %
         \node[const,left=of g] (a) {$\alpha_0$}; %
        % plate
         \plate [inner sep=.3cm,xshift=.02cm,yshift=.2cm] {plate1} {(x) (t) }{N}; %
         % \plate [inner sep=.3cm,xshift=.02cm,yshift=.2cm] {plate1} {(p) }{g}; %
        % edges
         \edge {y} {x}
         \edge {t} {x}
         \edge {g} {t}
         \edge {g0} {g}
         \edge {a} {g} }
  
      \caption{A representation of a Dirichlet process mixture model.}
\end{center}
\end{figure}
\vspace{-1\baselineskip}

\subsection{Hierarchical Dirichlet Processes}

\subsection{Contributions}
This paper makes the contributions as follows:
\begin{itemize}
\item We propose a new family of mixture models for unsupervised lexicon classification/Bayesian lexicon/BayesLex which can handle group clustering.
\item We provide a justification that aggregating lexicon scores is equivalent to a mixture of multinomial Na\"ive Bayes models.
\item We
\end{itemize}

\section{Related Works}
There are some works whose ideas have gone in the direction of incorporating Dirichlet processes to the Na\"ive Bayes classifiers. 

\section{Mixtures of Unsupervised Lexicon Classification}
\subsection{Mixture Models}
% Mixture models \cite{mclachlan1988mixture, mclachlan2019finite} are combinations of basic distributions which can give rises to model complex densities. A mixture model is usually in the form of a linear combination,
% \begin{equation}
% \begin{aligned}
%         f(y) &= \sum_{i=1}^{g} \pi_i f_i(y)\\
%         s.t.\  &\sum_{i=1}^{g} \pi_i = 1, \ 0 \le \pi_i \le 1,
% \end{aligned}    
% \end{equation}
% where $f$ denotes a mixture probability mass function, $f_i$ denotes a probability mass function of a base distribution which can also be called as a component of the mixture, $g$ is the number of finite mixtures (or the number of components), and $\pi_i$ is a mixing proportion or a mixing coefficient which also satisfies the requirements to be probabilities. Sometimes, we can view the mixing proportions as prior probabilities of picking the $i$th components. %For many base distributions, like Gaussians, maximizing the log-likelihood is intractable since the log of sums is hard to compute but for count-based distributions like Bernoulli, multinomial, or even priors like Beta or Dirichlet, the log-likelihood is the log of multiples which reduces to a more simple form.

Instead of modeling as a multinomial Na\"ive Bayes for the likelihood $P(x|y)$ in Eq. \ref{eq3}, we instead use a finite mixture of multinomial likelihood as,
\begin{equation}
\begin{aligned}
     \log{P(x|y)}& = \log{\sum_{i=1}^{g} \pi_i P_{multinomial_i} (x; \theta_y, N)}.\\
\end{aligned}
\end{equation}
This does not have a closed form solution since it is in the form of a sum inside a log function. Besides, it does not look like any aforementioned decision rules like Eq. \ref{eq1} or \ref{eq6} to justify its equivalent to lexicon classification. Therefore, we need to rearrange the terms using some identities. Here, we can use $(x + y) = x(1 + \frac{y}{x})$ and a logarithm product rule so that $\log{(x+y)} = \log{(x)} + \log{(1 + \frac{y}{x})}$. Then,
\begin{equation}
\begin{aligned}
     \log{P(x|y)}& = \log{\sum_{i=1}^{g} \pi_i P_{multinomial_i} (x; \theta_y, N)}\\
      & = \sum_{i=1}^{g} \log{ \pi_i P_{multinomial_i} (x; \theta_y, N)} \\
      &+ [ \log{( 1 + \frac{\sum_{j=2}^{g}{\pi_j P_{multinomial_j} (x; \theta_y, N)}} {\pi_1 P_{multinomial_1}(x; \theta_y, N)} )} \\ 
      &+ \log{( 1 + \frac{\sum_{j=3}^{g}{\pi_j P_{multinomial_j} (x; \theta_y, N)}} {\pi_2 P_{multinomial_2}(x; \theta_y, N)} )} \\ 
      &+ \dots \\
      &+ \log{( 1 + \frac{\pi_g P_{multinomial_g} (x; \theta_y, N)} {\pi_{g-1} P_{multinomial_{g-1}}(x; \theta_y, N)} )} ] \\
      & = \sum_{i=1}^{g} \log{ \pi_i P_{multinomial_i} (x; \theta_y, N)} + L(x, y).
\end{aligned}
\end{equation}
Applying this to the decision rule in Eq. \ref{eq2}, we can have,
\begin{equation}
\begin{aligned}
       \log{Pr(Y=0)} + \log{P(x|Y=0)} &\gtrless \log{Pr(Y=1)} + \log{P(x|Y=1)} \\
        \log{P(x|Y=0)} &\gtrless \log{P(x|Y=1)} \\
         \sum_{i\in W_0} [\sum_{j=1}^{g} x_i \log{\frac{ 1 + \gamma_j}{1 - \gamma_j}} + L(x, y)] &\gtrless  \sum_{i\in W_1} [\sum_{j=1}^{g} x_i \log{\frac{ 1 + \gamma_j}{1 - \gamma_j}}+ L(x, y)]. \\
\end{aligned}
\end{equation}
Clearly, the sums of log term obey the previous decision rule in Eq. \ref{eq6}. But, we still need to show that the $L(x, y)$ terms obey the decision rule too. Fortunately, by canceling out $\pi_j P_{multinomial_j}$ in the fractions of $L(x, y)$ terms, the remaining consists of only combinations of the $\pi$s and $\gamma$s, which we can further assume that are equal for both $Y$ and independent of $x$, so the $L(x, y)$ terms cancel out. We can have our new decision rule as a simple summation,
\begin{equation}
\begin{aligned}
 \sum_{i\in W_0} [\sum_{j=1}^{g} x_i \log{\frac{ 1 + \gamma_j}{1 - \gamma_j}}] &\gtrless  \sum_{i\in W_1} [\sum_{j=1}^{g} x_i \log{\frac{ 1 + \gamma_j}{1 - \gamma_j}}]. \\
\end{aligned}
\end{equation}
This is like aggregating scores from many lexicons and completes the proof for the finite mixture case. This is also consistent with some published empirical results, such as \cite{wiriyathammabhum2022promptshots, czarnek2022two}, whose results have shown that aggregating scores, even linearly or with different lexicon sets, improves the performance. Here, we provide a justification that aggregating many lexicons is equivalent to a finite mixture multinomial Na\"ive Bayes model. We also noted that the component base models must be different or else it is just no difference than a base single model. Furthermore, as can be easily seen in our proof, we fix the vocabulary of the lexicon, $W_0$ and $W_1$. Extending to aggregating many different lexicon sets might be straightforward. 

Moreover, another way to resolve this \cite{bishop2006pattern} is to introduce a latent variable $z$ which factorizes out the summation. 

\subsection{Dirichlet Process Mixtures}
Instead of modeling as a multinomial Na\"ive Bayes for the likelihood $P(x|y)$ in Eq. \ref{eq3}, we derive a new decision rule as we place a Dirichlet process prior on the multinomial Na\"ive Bayes to create a mixture of models using similar assumptions.
\begin{equation}
    P_{DP} (x;\alpha) = \int_{\pi} d\pi P(x|\pi)P(\pi|\alpha)
\end{equation}
\begin{equation}
\begin{aligned}
    \pi \ | \ \alpha_o \sim GEM(\alpha_0) &\qquad  z_i \ | \ \pi \sim \pi \\
    \phi_k \ | \ G_0 \sim G_0 &\qquad x_i \ | \ z_i, (\phi_k)_{k=1}^{\infty} \sim F(\phi_{z_i})
\end{aligned}
\end{equation}
\begin{equation}
    P_{DP} (x|y) = 
\end{equation}
From Eq. \ref{eq3}, we instead put the nonparametric Dirichlet process prior as,
\begin{equation}
\begin{aligned} 
    \log{P(x|y)}& = \log{P_{DP} (x;\alpha)}\\
    & = \log{P_{DP} (x;\alpha)}\\
\end{aligned}
\end{equation}

\subsection{Hierarchical Dirichlet Process Mixtures}

\section{Experimental Results}

\section{Conclusions and Discussions}

% \begin{credits}
% \subsubsection{\ackname} A bold run-in heading in small font size at the end of the paper is
% used for general acknowledgments, for example: This study was funded

% \subsubsection{\discintname}
% It is now necessary to declare any competing interests or to specifically
% state that the authors have no competing interests. Please place the
% statement with a bold run-in heading in small font size beneath the
% (optional) acknowledgments\footnote{If EquinOCS, our proceedings submission
% system, is used, then the disclaimer can be provided directly in the system.},
% for example: The authors have no competing interests to declare that are
% relevant to the content of this article. Or: Author A has received research
% grants from Company W. Author B has received a speaker honorarium from
% Company X and owns stock in Company Y. Author C is a member of committee Z.
% \end{credits}

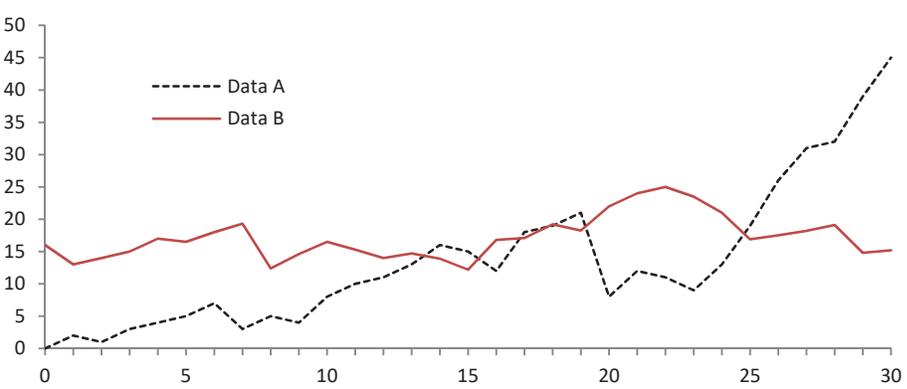
\end{document}